# Data-to-Text Bilingual Generation

Guy Lapalme
RALI-DIRO
`lapalme@iro.umontreal.ca`
November 24, 2023

**Abstract**

This document illustrates the use of _pyrealb_ for generating two _parallel_ texts (English and French) from a single source of data. The data selection and text organisation processes are shared between the two languages. only language dependent word and phrasing choices are distinct processes. The realized texts thus convey identical information in both languages without the risk of _being lost in translation_.  This is especially important in cases where strict and simultaneous bilingualism is required.  We first present the types of applications targeted by this approach and how the _pyrealb_ English and French realizer can be used for achieving this goal in a _natural_ way. We describe an object-oriented organization to ensure a convenient realization in both languages. To illustrate the process, different types of applications are then briefly sketched with links to the source code. A brief comparison of the text generation is given with the output of an instance of a GPT.

## 1. Contexts of application

Multilingual generation is important to be understood by a wider audience as shown by the always-increasing need for translation. In recent years, automatic translation has become an everyday tool for most people, especially non-English speakers, but its output must always be taken with care, especially when the target of the translation is not the mother tongue of the writer.  While humans are very good at filling missing information or correcting details in their language, automatic translations should always be revised by professionals for publication or official texts.

One difficult challenge in translation, both human and automatic, is ensuring that the information in the source and target texts are strictly equivalent, especially in the case of statistical data.  Although human translators work with great care, they do not always reproduce the numbers exactly in their translation which can be embarrassing and can have legal consequences. Rule-based and statistical automatic translators are less prone to these types of errors as they most often copy the values from the original to their translations. Although neural automatic translators produce very fluent texts, they are prone to _hallucinations_ because they start from an _abstraction_ of the original information, so their output must be checked carefully for ensuring that the same information is conveyed in both languages.

While automatic translation can be appropriate for texts written by humans, it is unnecessary when the text can be generated automatically in both languages. For example, in Canada, thousands of weather reports are generated daily from the output of numerical models.  Meteorologists use graphical tools to fine-tune the numerical outputs, but the English and French versions are generated automatically thus removing the translation delay and guaranteeing that the same information is conveyed in both languages as required by the Canadian Law. Similar arguments can be made for generating business reports for multinational corporations or sport narratives directly from data. Although in this report, we focus on data-to-text applications, we will show how bilingual generation can  also be used for creating translation drill exercises for students.

## 1.1. What is data-to-text ?

Before tackling the text generation process, we lay out the types of applications for which bilingual generation seems more appropriate. To simplify, we only consider the generation of a single sentence, but the process can be applied to all sentences of a text.

Adapting the notation introduced by Upadhyay and Massie (2022), we define a dataset $DB$ as a group of data instances called events $E_i$ each defined by a data structure $D_i$ for which a sentence $S$ must be generated for conveying the insights and the information about the event. The data structure $D_i$ is a set of objects $O_{ij}$ described by features $F_{ijk}$ for which values $R_{ijk}$ are recorded.

A text generator $g$ in a data-to-text context is thus a function from subsets of data structures to a sentence $S = g(\{D_i\})$.

Sometimes the $D_i$ are independent, for example, when describing a restaurant with a list of features such as the food, location, the prices, etc. But more realistically, time dependencies occur between events; for example, when describing sports matches in which it is important to convey relations between events to show the progress of a player or a team within a season. An important feature of this data set is thus some *time stamp*, such as a date or a link between games.

In this context, $g$ is *classically* divided in two subtasks (Reiter and Dale, 2000):

- $g_w$, (*What to say ?*) determines the content of the sentence by selecting the set of data structures to convey. This step being language independent, it is performed once for both languages.
- $g_{hl}$ (How to say ?) chooses the phrase structure and words to use in the sentence and performs the linguistic realization. This step, called *text realization*, is language dependent, thus the $l$ subscript. If target languages have some commonalities (such as between English and French), it is possible to share parts of the language dependent processing.

The generation of bilingual sentences $S_e$ and $S_f$ can thus be framed as function compositions where the selected events $Ds$ are identical for both languages.

$$Ds = g_w(\{D_i\})$$
$$S_e = g_{he}(Ds)$$
$$S_f = g_{hf}(Ds)$$

In *end-to-end generation* using neural methods that go directly from a data set, each event is represented as flattened tuples :

$$D_i = ((R_{i00}, R_{i01} \ldots R_{i0f}) \ldots (R_{in0}, R_{in1} \ldots R_{inf})) \quad (1)$$

from which the text is directly generated. It thus remains a challenge to ensure a consistent event selection for each language; moreover, there is always the risk of *hallucinations* (reporting events that do not appear in the original data) because of the generalization/abstraction process inherent in neural approaches.

We present how we use *pyrealb* to implement $g_{he}$ and $g_{hf}$ with a rule-based approach in different settings. Given that *pyrealb* is implemented in *python*, it can be conveniently combined with data-processing steps $g_w$ implemented using one of the many *python* data analysis tools. In our examples, we use simple *python* functions for implementing the data selection and performing common linguistic choices.

## 2. Multilingual realizers

A number *multilingual text generators* (i.e. dealing with at least another language than English) have been developed. For example, [KPML](#) can handle Spanish, Dutch, Chinese, German and Czech; [Surgeon-2](#) can generate German; [Grammatical Framework](#) (Ranta 2011) is a programming language designed for writing grammars in several languages in parallel; GenDR (Lareau et al. 2018) can generate sentences in Catalan, French, Polish, Portuguese and Spanish. These generators are based on linguistic theories, considering many details in the construction of sentences, which allows powerful realizations. However, that complexity hinders somewhat their ease of use: writing specifications for them requires an intimate knowledge of the underlying theory.

[SimpleNLG](#) (Gatt and Reiter 2009), as its name implies, defines itself by its ease of learning and of use. Words, phrases and other structures are Java objects created and manipulated by a programmer and integrated into a Java project. SimpleNLG can also be called from other programming languages through a web server with an XML interface. While its principles somewhat limit the power of its realizations compared to other systems, these realizations are adequate for many uses. It has been ported to some languages, namely [Galician](#), [Spanish](#), [German](#), [Dutch](#), [Italian](#) and [Mandarin](#) but a single language at a time, one exception being [SimpleNLG-EnFr](#) that Vaudry and Lapalme developed to work in both English and French at the same time.

Building on this experience, we developed [*jsRealB*](#) (Lapalme, 2022), written in JavaScript, to ease its integration in a web environment. [RosaeNLG](#) is a Natural Language Generation library for *node.js* or browser execution, based on the [Pug template engine](#) dealing with English, French, German, Italian and Spanish. RosaeNLG was developed for realizing some simple data to text applications and is especially tuned for outputting lists of objects and properties using appropriate commas and a conjunction at the end of the list. Its linguistic coverage, at least for French and English, is limited compared to *jsRealB*.

We later ported *jsRealB* to *python* to create [*pyrealb*](#), described in the next section, with the same goal of realizing sentences in both English and French, even within the same sentence. The further sections will show how *pyrealb* can integrate all steps ($g_w$, $g_{he}$ and $g_{hf}$) of the data-to-text pipeline in a single and convenient *python* formalism.

## 3. *pyrealb*

[*pyrealb*](#) is a Python package which allows English and French sentence realization by programming language instructions that create internal data structures corresponding to the elements of the sentence. The data structure can be built incrementally and, when needed, the realization process traverses it to produce a string in the appropriate language.

*pyrealb* has the following components for both English and French:

- a *lexicon* (a *JSON* file) defining the word category, gender, number, declension and conjugation rule number and other features needed to produce the final token;
- *morphological rules* (a *JSON* file) to determine the appropriate word forms, such as plurals and conjugations;
- *syntactic rules* (*python* classes) to build sentence structures from terminals and properly order words within a sentence, performing the most common agreements between constituents and carrying out other useful sentence organization tasks such as managing coordination or applying sentence transformations.

*pyrealb* also performs the spelling out of numbers and the wording of temporal expressions that are especially useful in data to text applications.

*pyrealb* accepts either a *Constituent* or a *Dependent* notation for building sentences. In this document, we give examples using the *Constituent* notation, but the same methodology applies to the *Dependent* notation. The data structure is built by class constructor calls whose names are like the symbols typically used for constituent syntax trees:

- **Terminal**: `N` (noun), `V` (verb), `A` (adjective), `D` (determiner), `Pro` (pronoun), Adv (adverb), `P` (preposition), `C` (conjunction), `NO` (number), `DT` (date), `Q` (quoted/canned text). A terminal is created with a single parameter its lemma, most often a string,
- **Phrase** for combining its parameters, i.e. terminals and other phrases: `S` (Sentence), `SP` (Subordinate Phrase), `NP` (Noun Phrase), `VP` (Verb Phrase), `AP` (Adjective Phrase), `CP` (Coordinate Phrase), `PP` (Prepositional Phrase), `AdvP` (Adverbial Phrase).

To produce the text string corresponding to the structure of a **Terminal** or a **Phrase**, the `realize()` method of **Terminal**, but more often of a **Phrase**, must be called. As most often generation occurs in only one language at a time, *pyrealb* tracks the current language set with either `loadEn()` or `loadFr()` after which terminals and phrases created are associated with this language. Each **Terminal** being associated with a language when it is created, the appropriate morphological rules can be applied when it is realized.

Features are added to these structures using the dot notation to modify their properties. For terminals, their person, number, gender can be specified. For phrases, the sentence may be negated or set to a passive mode; a noun phrase can be pronominalized. Punctuation signs and HTML tags can also be added.

*pyrealb* deals with the *final realization*, which is an often neglected part in NLG systems because it is dubbed to be pedestrian, often associated with glorified format statements, although its output is the only thing that the end user sees. How acceptable is an output if word agreements or elision are not properly done or if it consists of a mere list of tokens? This might be sufficient for automatic evaluation, but it cannot be used in a production setting. A well formatted and grammatically correct output is important for the social acceptability of a system. The fact that neural systems often produce *flabbergastingly* fluent text explains in part their popularity.

Because English and French share most of their grammatical features, options can be specified for both languages, except for a few cases; e.g. English perfect aspect is ignored in French and French tenses *imparfait* and *temps composés* are not used in English.

The following shows how an English and a French sentence can be built and printed. Note that adjectives are placed according to the rules of each language.

```
loadEn()                                                  # set the language to English
print(S(NP(D("the"),N("cat"),A("small")),                 # create a subject NP
        VP(V("jump").t("ps"),                             # create VP, setting past for the verb time
           PP(P("on"),                                    # create a PP with
              NP(D("the"),N("mat"),A("green"))))          #    an object NP
        ).realize())
 # output: The small cat jumped on the green mat.

loadFr()                                                  # set the language to French
print(S(NP(D("le"),N("chat"),A("petit")),                 # create a subject NP
        VP(V("sauter").t("ps"),                           # create VP, setting past for the verb time
           PP(P("sur"),                                   # create a PP  with
              NP(D("le"),N("tapis"),A("vert"))))          #    an object NP
        ).realize())
 # output: Le petit chat sauta sur le tapis vert.
```

It is also possible to mix languages within a single sentence such as in the following French sentence with an English subject. Note that the plural of the English subject is propagated to the French portion of the sentence. In practice, this type of bilingual sentence is seldom used, but it was thought important to cater also for these cases.

```
loadEn()                                                  # set the language to English
subj = NP(D("the"),N("cat"),A("small")).n("p")            # create an English plural NP
loadFr()                                                  # set the language to French
verb = VP(V("sauter"),                                    # create a French VP, present by default
          PP(P("sur"),                                    # create a PP with
             NP(D("le"), N("tapis"), A("vert"))))         # an object NP
print(S(subj,
        verb).realize())
 # output: The small cats sautent sur le tapis vert.
```

*pyrealb* «walks the talk» by calling *itself* for realizing its error messages in the *current* language such as missing words from the lexicon or bad values for options. English is used for errors detected in English sentences and similarly for French.

## 3.1. Useful tools for data-to-text generation

Data being unpredictable, it is often hard to create a complete *pyrealb* expression with all its components in a single call. So *pyrealb* allows an incremental way of building the structure using the `add(elem,pos)` method to modify an existing **Phrase** by adding a new parameter at a position (last by default). The following example adds a complement to the verb phrase of the previous example.

```
1  loadEn()
2  verb.add(PP(P("over"),NP(D("a"),N("fence")).n("p")))
3  print(S(subj.n("p"),   # set the English subject to plural
4          verb).realize())
5   # output: The small cats sautent sur le tapis vert over fences.
```

Realizing a variable number of data is also critical in a data-to-text context and within a sentence this implies building the coordination of elements. The following shows how coordination adapts its realization to the number of arguments. Note also the fact that **Phrase** constructors accept lists of parameters that are *flattened* before construction of the data structure.

```
1  persons = ["mother","daughter","father"]
2  for i in range(0,len(persons)):
3      print(S(CP(C("and"),[NP(D("the"),N(p)) for p in persons[:i+1]]),
4              VP(V("be"),
5                 A("happy"))).realize())
6  # output:
7  # The mother is happy.
8  # The mother and the daughter are happy.
9  # The mother, the daughter and the father are happy.
```

We see that the coordination is ignored when there is only one element and that a comma is introduced when there are more than two. The number for the verb depends on how many elements are coordinated. The following example shows a similar case in French, in which the gender and number of both the verb and adjective depend on the number of coordinated subjects according to grammatical rules of French.

```
1   loadFr();
2   personnes = ["mère","fille","père"]
3   for i in range(0, len(personnes)):
4       print(S(CP(C("et"), [NP(D("le"), N(p)) for p in personnes[:i + 1]]),
5               VP(V("être"),
6                  A("heureux"))).realize())
7    # output:
8    # La mère est heureuse.
9    # La mère et la fille sont heureuses.
10   # La mère, la fille et le père sont heureux.
```

In any *python* program, functions can/should be defined for creating recurrent patterns such as the following which creates a sentence structure for reporting an event involving some persons at a date. The tense of the verb can also be specified.

```
from datetime import datetime
loadEn()
def report(event, persons, date, tense="p"):
    meeting = PP(P("at"), NP(D("a"), N(event)))
    return S(CP(C("and"), [NP(D("a"),N(person)) for person in persons]),
             NP(NO(len(persons)),N("person")).ba("("), # show number of persons
             VP(V("be").t(tense),
                A("present"),
                meeting,
                DT(date).dOpt({"hour":False,"minute":False,"second":False})))

print(report("birthday",["mother","girl"],
             datetime(2023,5,30),"ps").realize())
print(report("assembly",["grandfather","father","boy"],
             datetime(2023, 12, 30),"f").realize())
 # output:
 # A mother and a girl (2 persons) were present at a birthday on Tuesday, May 30, 2023.
 # A grandfather, a father and a boy (3 persons) will be present at an assembly on Saturday, December 30, 2023.
```

## 3.2. Challenges for bilingual generation

Because, in a bilingual setting, there are two language contexts with their own rules, care must be given at the evaluation time of the expression. In *python*, all *top-level* expressions in scripts such as the ones shown above are evaluated when the script is loaded, it is important to set the appropriate language environment (using `loadEn()` or `loadFr()`) before they were encountered and evaluated.

To defer the evaluation, it is possible to use a function (`def` in *python*) whose body will be evaluated when it is called. As *python* uses the *applicative order* evaluation mechanism, the parameters of a function are evaluated before its call, so the appropriate language context must also be set when the function is called. To delay the evaluation of a python expression `exp` until it is needed, `lambda:` can be added before `exp` which creates a function whose body can be later evaluated by calling `exp()`. As we saw earlier, this brings the advantage that this function creates new copies of the original structure. Parameters can also be added to the lambda for more flexibility.

Although not specific to bilingual generation, delaying expression evaluation is also useful in the context of the `oneOf(...)` function, which selects radnomly one of its arguments. `oneOf(...)` is particularly useful for varying between synonyms or equivalent phrasings to make the text less repetitive. `oneOf(...)` checks if the selected element is callable, and if so it calls it and returns the result of this evaluation. So `oneOf()` is often called with functional parameters such as these:

```
1   oneOf(lambda: expr_1,
2         lambda: expr_2,
3         ...
4         lambda: expr_n)
```

Without `lambda`, given the applicative order of evaluation of `python`, all `expr_i` would be evaluated but return only one of them.

We have illustrated some features of the *pyrealb* realizer. For more details, see the [online documentation](#) or experiment with a [Jupyter Notebook](#).

The next sections give examples of bilingual text generation in data-to-text contexts. Most of the data processing and text organization is common to both languages, the only language-specific part being the final realization. This setup thus greatly simplifies ensuring that the same information is conveyed in both languages.

# 4. Organizing the realization process with *pyrealb*

As a first example of a bilingual report, we consider the case where there is a strict parallelism between English and French: only words differ, the phrase structure is identical for both languages. This simplification will be removed later, but it allows focusing on some aspects.

## 4.1. Common phrase structure

Names of persons are added as nouns to each lexicon and a series of equivalent words in English and French are given. A function is defined for determining the appropriate tense to use.

```
1   # add names to the English and French lexica
2   for loadF in [loadEn,loadFr]:
3       loadF()
4       addToLexicon({"Alice":{ "N": {"g": "f", "tab": "nI" } }})
5       addToLexicon({"Bob":{ "N": {"g": "m", "tab": "nI" } }})
6       addToLexicon({"Eve":{ "N": {"g": "f", "tab": "nI" } }})
7
8   # text parameterization with words dictionaries indexed by language
9   participants = ["Alice", "Eve", "Bob"]
10  conj        = {"en":"and",     "fr":"et"}
11  prep        = {"en":"at",      "fr":"à"}
12  det         = {"en":"a",       "fr":"un"}
13  copula      = {"en":"be",      "fr":"être"}
14  attribute   = {"en":"present", "fr":"présent"}
15  individual  = {"en":"person",  "fr":"personne"}
16  dateOptions = {"minute":False,"second":False}
17
18  # compare day of date with the day of the reference
```

```python
19  def tense(date, reference):
20      o = date.toordinal()
21      ref_o = reference.toordinal()
22      return "p" if o == ref_o else "f" if o > ref_o else "ps"
```

The realization function is like the `report` function in the previous section, the main difference being that words are indexed by the `lang` parameter.

```python
1  def report(event, persons, date, lang):
2      loadEn() if lang=="en" else loadFr()
3      meeting = PP(P(prep[lang]), NP(D(det[lang]),N(event)))
4      return S(CP(C(conj[lang]), [N(person) for person in persons]),
5                 NP(NO(len(persons)).nat(), N(individual[lang])).ba("("),
6                 VP(V(copula[lang]),
7                    A(attribute[lang]),
8                    meeting,
9                    DT(date).dOpt(dateOptions)).t(tense(date,today))))
```

This function can be called to create sentences in both French and English, varying the number of participants and the date.

```python
1  today = datetime.today()
2  loadEn(); print(DT(today).dOpt(dateOptions).realize(),end="")
3  loadFr();print("-",DT(today,"fr").dOpt(dateOptions).realize(),"\n")
4  for (i,day) in zip(range(1,len(participants)+1),
5                     [today-timedelta(days=1),today,today+timedelta(days=1)]):
6      print(report("assembly",participants[:i], day,"en").realize())
7      print(report("réunion",participants[:i], day,"fr").realize())
8      print("--")
9
```

and produces the following bilingual output, in which dates and numbers are properly written with the correct agreements between components although the user did not specify them explicitly.

```
 1  on Tuesday, September 26, 2023 at 5 p.m.- le mardi 26 septembre 2023 à 17 h
 2
 3  Alice (one person) was present at an assembly on Monday, September 25, 2023 at 5 p.m.
 4  Alice (une personne) fut présente à une réunion le lundi 25 septembre 2023 à 17 h.
 5  --
 6  Alice and Eve (two persons) are present at an assembly on Tuesday, September 26, 2023
    at 5 p.m.
 7  Alice et Eve (deux personnes) sont présentes à une réunion le mardi 26 septembre 2023
    à 17 h.
 8  --
 9  Alice, Eve and Bob (three persons) will be present at an assembly on Wednesday,
    September 27, 2023 at 5 p.m.
10  Alice, Eve et Bob (trois personnes) seront présents à une réunion le mercredi 27
    septembre 2023 à 17 h.
11  --
```

## 4.2. Different phrase structures

The example in the previous section is admittedly restrictive, because it takes for granted that the phrase structure in both languages is identical. This is like localization tools used to adapt computer applications to different languages by adapting menu items and user messages. But this approach cannot always be used in more realistic text generation contexts.

We will now show a way to generate similar sentences in both languages while keeping some flexibility in the formulations using an object-oriented organization. The language independent algorithms and phrase choices are performed in a class and the language dependent parts are done in subclasses. Usually the subclasses have a similar organization but they allow different sentence structures.

Here is the language independent main class equivalent to our previous example. Word choices will be performed in the subclasses attribute and methods such as `self.and_conj`, `self.attend()` or `self.meeting()`.

```
 1  class Realizer:
 2      def __init__(self): # called by __init__() in subclasses after setting the language
 3          addToLexicon({"Alice":{ "N": {"g": "f", "tab": "nI" } }})
 4          addToLexicon({"Bob":{ "N": {"g": "m", "tab": "nI" } }})
 5          addToLexicon({"Eve":{ "N": {"g": "f", "tab": "nI" } }})
 6
 7          today      = datetime.today()
 8          dateOptions = {"minute": False, "second": False}
 9
10          # compare day of date with the day of the reference
11      def tense(self,date, reference):
12          o = date.toordinal()
```

```
13            ref_o = reference.toordinal()
14            return "p" if o == ref_o else "f" if o > ref_o else "ps"
15
16        def report(self,event,persons,date):
17            print(S(CP(self.and_conj,[N(person) for person in persons]),
18                    NP(NO(len(persons)).nat(), self.individual()).ba("("),
19                    self.attend(self.meeting(event)),
20                    DT(date).dOpt(Realizer.dateOptions))
21                        .t(self.tense(date,Realizer.today)))
```

The language-specific parts are the following `English` and `Francais` classes. The `report()` method is also defined in each subclass to that the appropriate language is *loaded* before calling the language independent part is called via `super()`. The terminals are specified directly in each language. Note that the sentence structure for `attend(meeting)` is different in subclasses.

```
1   class English(Realizer):                        |   class Francais(Realizer):
2     def __init__(self):                           |     def __init__(self):
3         loadEn()                                  |         loadFr()
4         self.and_conj = C("and")                  |         self.and_conj = C("et")
5         super().__init__()                        |         super().__init__()
6                                                   |
7     def report(self,event,persons,date):          |     def report(self,event,persons,date):
8         loadEn()                                  |         loadFr()
9         super().report(event,persons,date)        |         super().report(event,persons,date)
10                                                  |
11    def attend(self,meeting):                     |     def attend(self,meeting):
12        return VP(V("attend"),meeting)            |         return VP(V("être"),A("présent"),
13                                                  |                   PP(P("à"),meeting))
14                                                  |
15    def individual(self):                         |     def individual(self):
16        return N("person")                        |         return N("individu")
17                                                  |
18    def meeting(self,noun):                       |     def meeting(self,noun):
19        return NP(D("the"), N(noun))              |         return NP(D("le"), N(noun))
20
```

These language dependent classes are first instantiated, then called as follows

```
1   english = English()
2   francais = Francais()
3   for (i,day) in zip(range(1,len(participants)+1),
4                      [today-timedelta(days=1),today,today+timedelta(days=1)]):
5       english.report("assembly",participants[:i],day)
6       francais.report("réunion",participants[:i],day)
```

To get a similar output as the previous example, except for the date and the way of indicating attendance.

```
 1  on Friday, September 29, 2023 at 2 p.m.- le vendredi 29 septembre 2023 à 14 h
 2
 3  Alice (one person) attended the assembly on Thursday, September 28, 2023 at 2 p.m.
 4  Alice (un individu) fut présente à la réunion le jeudi 28 septembre 2023 à 14 h.
 5  --
 6  Alice and Eve (two persons) attend the assembly on Friday, September 29, 2023 at 2
    p.m.
 7  Alice et Eve (deux individus) sont présentes à la réunion le vendredi 29 septembre
    2023 à 14 h.
 8  --
 9  Alice, Eve and Bob (three persons) will attend the assembly on Saturday, September 30,
    2023 at 2 p.m.
10  Alice, Eve et Bob (trois individus) seront présents à la réunion le samedi 30
    septembre 2023 à 14 h.
```

This setup for a single sentence shows how the object-oriented features of Python can be used to organize the parallel sentence realization in two languages. The full code is available on GitHub as demo for pyrealb. In such a simple case, the class organization might seem an *overkill*, but this organization is very convenient in more complex cases as it will be shown later.  When the realization process makes use of *Abstract Base Classes*, the python interpreter can check that the language dependent realizer methods are similar in all subclasses and help guarantee that equivalent information is conveyed in both languages, provided each method with the same name provide equivalent phrasings.

The next section describes use cases for bilingual data-to-text realization in more complex settings, but the fundamental idea is the same: parallel syntactic abstractions in a convenient notation that can be parameterized with values.  This ensures that the input data is correctly conveyed in the output thus removing the need for double-checking or having to install guardrails to avoid *hallucinations* (reporting facts that are not present in the data) or risking the output of inappropriate language.

When the situation is appropriate, namely, when we deal with numerical data, we must be certain that computer systems always produce the right answer, not just *usually*. Remember the *Pentium bug* that affected *1 in 9 billion floating point divides* but that cost Intel 475 million in 1994.

As Martin Kay (1980) put it

> An algorithm that works most of the time is, in fact, of very little use unless there is some automatic way of deciding when it is and when it is not working.

# 5.  Use cases

This section gives data-to-text demonstration programs implementing our methodology. As the full code and some algorithmic details are available on the pyrealb GitHub, we focus on the specificities of the data of each application and display typical outputs.

## 5.1.  Realization of all the data

We now present two use cases in which the input data has already been selected and the generation process is limited to the presentation of all the data. The generation is thus limited to *How to say ?* ($g_{he}$ and $g_{hf}$) which can imply sorting and organizing the input data, though.

### 5.1.1.  E2E challenge

[code]

The task here is to realize descriptions of restaurants based on a meaning-representation given by a list of key-value pairs such as the following;

```
1  food[English], priceRange[high], near[Raja Indian Cuisine],
2   name[The Mill], area[riverside], familyFriendly[yes], eatType[pub]
```

About 50K pairs of meaning-representation with the corresponding expected text were crowdsourced for a shared task, held at the 2017 SIGdial meeting (Dušek *et al.*, 2020). We ported to *python* our previous jsRealB version, described in this page. This system produces the following two sentences from the meaning-representation given above.

```
1  The Mill is a pub near Raja Indian Cuisine in the riverside area that serves English
2  food with high prices. It is kid friendly.
3
4  The Mill est un pub près de Raja Indian Cuisine au bord de la rivière qui offre une
5  cuisine anglaise à prix élevés. Il est approprié pour les enfants.
```

### 5.1.2.  WebNLG Challenge 2020

[code]

This task is to realize information given as *simplified* RDF triples. An RDF triple is composed of three URIs corresponding to the subject, the predicate and the object that can also be a constant string, a date or a number. The predicate of a triple declares a relation between the subject and the object, such as `Alan_Bean | birthPlace | Wheeler,_Texas`, in which `Alan_Bean` is the subject, `birthPlace` the predicate indicating that the subject was born at the place given by the object and `Wheeler,_Texas` is the object. This could be verbalized as *Alan Bean is born in Wheeler, Texas.*

```
1  Apollo_12  | backupPilot   | Alfred_Worden
2  Alan_Bean  | mission       | Apollo_12
3  Apollo_12  | operator      | NASA
4  Apollo_12  | commander     | David_Scott
5  Alan_Bean  | birthPlace    | Wheeler,_Texas
6  Alan_Bean  | selectedByNasa | 1963
7  Alan_Bean  | birthDate     | "1932-03-15"
```

The English RDF verbalizer is based on a symbolic approach: each RDF triple corresponds to a sentence in which the subject and the object of a triple are mapped almost verbatim as subject and object of the sentence. This is possible in this case because the subject and object have already been *nominalized*, but that would not be the case if *real* URIs had been inputted.

The predicate of the triple corresponds to a verb phrase which determines the structure of the sentence. The predicates are ordered to create a meaningful story and parts of sentences are merged when they share subjects or predicates. [Our participation at the WebNLG Challenge](#) used *jsRealB*, through a web server, for the English realization. This system obtained good evaluation results (being in the middle of the pack) for automatic evaluation. For the human evaluation, it was judged excellent (always in the first group of participants) for coverage, relevance and correctness. The text structure and fluency were judged less well (in the second and third group).

The current version uses *pyrealb* and realizes both English and French sentences. The following sentences were realized from the input given above.

```
1  Alan Bean was born on March 15, 1932 in Wheeler, Texas and joined NASA in 1963.
2  He was a crew member of Apollo 12. Apollo 12 is commanded by David Scott, it has as
3  its back-up pilot Alfred Worden and is operated by NASA.
4
5  Alan Bean est né le 15 mars 1932 à Wheeler, Texas et a été choisi par NASA en 1963.
6  Il a été un membre de l'équipe d'Apollo 12.  Apollo 12 est commandé par David Scott, il
7  a compté Alfred Worden comme pilote de réserve et est opéré par NASA.
```

[More details](#)

## 5.2.  Realization of a subset of the data

We now show examples of cases with *plenty of data*, for which to generate bilingual texts that focus on some important aspects.

### 5.2.1.  Weather reports

[[code](#)]

The input of the application is a set of meteorological information (e.g., precipitations, temperature, wind, UV index, ...) provided by [Environment and Climate Change Canada](#) (ECCC). Unlike many data-to-text applications, this information is machine generated: it is created by a numerical weather model which outputs data for ranges of hours after the time the bulletin is expected to be issued.

For this demo, we extracted a subset of the global information for regions of Ontario and Québec for 2018 and 2019 which is nevertheless illustrative of the natural language generation problems encountered in this context. We converted the Meteocode, an internal data format of ECCC, to JSON in which time indications are *shifted*, so that they appear in local time while, in the original, they were in UTC.

We now outline the JSON data organization for a weather bulletin used as input for our demonstration program in terms of Python data structures:

- administrative information: issue and next-issue times, list of region names to which the forecast applies in both English and French
- weather information : list of values of which the first two are the *starting hour* and *ending hour* relative to 0h of the issue datetime, when they are negative, they refer to historical data; the other values (described below depending on the type of information). For `precipitation-type` and `wind`, a value can be a list of values which describes an exceptional phenomenon (e.g., *gust* within a wind period) that occurs during this period.

For a given period, these JSON terms can be visualized as follows:

| **tomorrow** | **( 6h,18h) fpto12-2018-07-18-2000-r1209c :: 2018-07-18 16:00:00** |
| --- | --- |
| precipitation-type | [15h,0h):[showers, [15h,0h):[thunderstorm]] |
| precipitation-probability | [5h,15h):[10], [15h,18h):[30] |
| sky-cover | [5h,11h):[2, 2], [11h,15h):[2, 8], [15h,18h):[8, 8] |
| temperatures : | [5h,8h):[15], [8h,11h):[23], [11h,14h):[28], [14h,17h):[25], [17h,20h):[23] |
| uv-index | [12h,14h):[7.7] |
| wind | [0h,12h):[sw, speed, 10], [12h,20h):[sw, speed, 20] |

Although in principle, weather data is strongly time-dependent, the upstream process ensures that the necessary historical information is included in the current record. Thus $g_w$ limits itself to the selection of the most important values within the current dataset selection according to standard values defined by writing rules of Environment Canada.

Here is an example of an *evening* bulletin realized by **pyrealb** in English and French.

```
 1  WEATHER BULLETIN: regular
 2  Forecasts issued by pyrealb on Wednesday, July 18, 2018 at 4:00 p.m. for today and
    tomorrow at 4:00:00 p.m.
 3  The next scheduled forecasts will be issued on Thursday, July 19, 2018 at 5:30 a.m.
 4  Armstrong - Auden - Wabakimi Park
 5  Nakina - Aroland - Pagwa
 6
 7  Tonight : Clear.  A few clouds.  Partly cloudy.  30 percent chance of
 8    showers.  Wind west 20 km/h around noon.  Becoming southwest in the
 9    evening.  Low 14, with temperature rising to 28 by morning.
10  Thursday : Mainly sunny.  Increasing cloudiness tomorrow morning.
11    Mainly cloudy.  30 percent chance of showers.  Wind southwest 20 km/h
12    around noon.  High 28.  Low 15.  UV index 8 or very high.
13  Thursday night : Mainly cloudy.  30 percent chance of showers.  Wind
14    southwest 20 km/h around noon.  Low 14, with temperature rising to 23
15    by morning.
```

```
16    END
17
18    BULLETIN MÉTÉOROLOGIQUE: régulier
19    Prévisions émises par pyrealb le mercredi 18 juillet 2018 à 16 h 0 pour aujourd'hui et
      demain à 16 h 0 min 0 s.
20    Les prochaines prévisions seront émises le jeudi 19 juillet 2018 à 5 h 30.
21    Armstrong - Auden - parc Wabakimi
22    Nakina - Aroland - Pagwa
23
24    Ce soir et cette nuit : Dégagé.  Quelques nuages.  Partiellement
25     couvert.  30 pour cent de probabilité d'averses.  Vents de l'ouest de
26     20 km/h vers midi.  Devenant du sud-ouest dans la soirée.  Minimum
27     14, températures à la hausse pour atteindre 28 en matinée.
28    Jeudi : Généralement ensoleillé.  Ennuagement demain matin.
29     Généralement nuageux.  30 pour cent de probabilité d'averses.  Vents
30     du sud-ouest de 20 km/h vers midi.  Maximum 28.  Minimum 15.  Indice
31     UV 8 ou très élevé.
32    Jeudi soir et nuit : Généralement nuageux.  30 pour cent de
33     probabilité d'averses.  Vents du sud-ouest de 20 km/h vers midi.
34     Minimum 14, températures à la hausse pour atteindre 23 en matinée.
35    FIN
```

As the bilingual outputs for both French and English are strictly parallel and a complete bulletin is generated in a language, $g_h$ uses parallel bilingual structures within the code such as the following to determine the phrase structure for the day depending on the hour (e.g., `morning` or `matin` when the hour is between 9 and 12). Most generation function are parameterized by the language to generate.

```
1   dayPeriods=[(0,5, {"en":lambda:NP(N("night")),
2                      "fr":lambda:NP(N("nuit"))}),
3               (5,9, {"en":lambda:NP(Adv("early"),N("morning")),
4                      "fr":lambda:NP(N("début"),PP(P("de"),N("matinée")))}),
5               (9,12, {"en":lambda:NP(N("morning")),
6                       "fr":lambda:NP(N("matin"))}),
7               (12,18,{"en":lambda:NP(N("afternoon")),
8                       "fr":lambda:NP(N("après-midi"))}),
9               (18,24,{"en":lambda:NP(N("tonight")),
10                      "fr":lambda:NP(N("soir"))})]
```

[more details](more details)

## 5.2.2. Basketball summaries

[code]

We now present a case of time-dependent data used for generating English and French *statistic-focused summaries of basketball games* using information found in the SportSett:Basketball dataset (Thomson *et al.*,2020). This dataset combines scores and performance measures about the teams and the players of thousands of NBA games with human-authored summaries about these games.

The detailed statistics give information about the number of points, of attempted and made field goals, of blocks, of assists, etc. The following table gives the box scores for the Philadelphia 76ers in their game against the Miami Heat on November 1, 2014, the first game of the *Train* dataset which we use as a running example in this paper. The heads of tables follow the same conventions as the one used for the official scores.

| Game | FGM | FGA | FG3M | FG3A | FTM | FTA | OREB | TREB | AST | STL | BLK | TOV | PF | PTS |
|---|---|---|---|---|---|---|---|---|---|---|---|---|---|---|
| Q1 | 13 | 21 | 3 | 8 | 1 | 2 | 2 | 10 | 10 | 2 | 5 | 6 | | 30 |
| Q2 | 8 | 15 | 0 | 2 | 8 | 12 | 1 | 11 | 7 | 1 | 1 | 4 | | 24 |
| Q3 | 10 | 16 | 4 | 8 | 5 | 6 | 1 | 10 | 9 | 2 | 2 | 6 | | 29 |
| Q4 | 4 | 15 | 0 | 5 | 5 | 6 | 0 | 6 | 2 | 4 | 2 | 8 | | 13 |
| game | 35 | 67 | 7 | 23 | 19 | 26 | 4 | 37 | 28 | 9 | 10 | 24 | 21 | 96 |

The following shows the data for the 4 (out of the 12) Philadephia players scoring the most points in this game.

| PLAYER | STRT | MIN | FGM | FGA | FG3M | FG3A | FTM | FTA | OREB | TREB | AST | STL | BLK | TOV | PF | PTS | +/- |
|---|---|---|---|---|---|---|---|---|---|---|---|---|---|---|---|---|---|
| Tony Wroten | True | 33 | 6 | 11 | 1 | 4 | 8 | 11 | 0 | 3 | 10 | 1 | 1 | 4 | 1 | 21 | -11 |
| Brandon Davies | False | 23 | 7 | 9 | 1 | 2 | 3 | 4 | 0 | 3 | 0 | 3 | 0 | 3 | 3 | 18 | -1 |
| Hollis Thompson | True | 32 | 4 | 8 | 2 | 5 | 0 | 0 | 0 | 1 | 2 | 0 | 3 | 2 | 2 | 10 | -17 |
| Henry Sims | True | 27 | 4 | 9 | 0 | 0 | 1 | 2 | 1 | 4 | 2 | 0 | 1 | 0 | 1 | 9 | -10 |

In this case, the summary reports information about the team and the players in the current game. This implies statistical procedures for determining the winners, the best players and showing turning points and important differences about some aspects of the game (e.g. field goals, three-pointers, etc.) between teams in each quarter.

Basketball game summaries also consider information from previous games to identify winning or losing streaks or to mention that a performance is above average in the season. Season statistics are also used to identify the outstanding players (about three or four out of more than twenty).

This is an example where data selection $g_w$ must consider not only the current $D_i$ but also others either the past games of the season or even information gathered about the past seasons. The result $D_s$ of this selection process is performed once and is used for both $g_{he}$ and $g_{hf}$.

Here is English summary produced by the system from the above data:

> The Heat (2-0) , leader in their conference, defeated the 76ers (0-3) 114-96 at the Wells Fargo Center in Philadelphia on Saturday.

> The Heat led in all four quarters. Over the first quarter, the 76ers obtained better goals percentage, a difference of 14%. Over the third quarter, the 76ers got better free throws percentage, an advantage of 21%. The Heat dominated the 76ers for points by 14 over the fourth quarter. In the game, the Heat obtained better three-pointers percentage, 50% to 30%.
>
> Chris Bosh who was a starter led the way, posting 30 points (9-17 FG, 2-5 3Pt, 10-11 FT) while adding four assists. Tony Wroten who started this game scored a game high with 21 points (6-11 FG, 1-4 3Pt, 8-11 FT) and ten assists and performed a double-double.
>
> The Heat showed 49 percent from the field and 20-29 free throws.  Mario Chalmers who was a starter added 20 points.  Luol Deng contributed 15 points with 7-for-11 FG.  Shawne Williams added 15 points (5-9 FG, 3-5 3Pt, 2-2 FT) while adding four assists.  Dwyane Wade had nine points (4-18 FG, 0-1 3Pt, 1-3 FT) while adding ten assists.
>
> The 76ers showed 52 percent from the field and 19-26 attempts at the charity stripe and committed 24 turnovers.  Brandon Davies who was a starter ended up with 18 points with 7-for-9 FG.  Luc Mbah a Moute recorded nine points with seven rebounds grabbed and three assists.  Malcolm Thomas had eight points in 19 minutes.  Alexey Shved posted six points (1-4 FG, 1-4 3Pt, 3-3 FT) while adding six assists.
>
> The Heat' next game will be at home against the Toronto Raptors on Sunday. The 76ers' next game will be at home against the Houston Rockets on Monday

The summary in French is the following

> Le Heat (2-0) , meneurs dans leur conférence, a dominé les 76ers (0-3) 114-96 au stade Wells Fargo Center samedi à Philadelphia.
>
> Le Heat a mené pendant les quatre quarts. Durant le premier quart, les 76ers ont réussi les meilleurs lancers en pourcentage, une différence de 14%. Pendant le troisième quart, les 76ers ont réussi les meilleurs lancers francs en pourcentage, un avantage de 21%. Le Heat a dominé les 76ers pour les points par 14 pendant le quatrième quart. Durant la partie, le Heat a obtenu les meilleurs tirs à 3 points en pourcentage, 50% en comparaison avec 30%.
>
> Chris Bosh qui débutait la partie a réalisé une performance excellente comptant 30 points (9-17 L, 2-5 L3, 10-11 LF) et quatre passes décisives. Tony Wroten qui débutait la partie a obtenu le meilleur pointage du match avec 21 points (6-11 L, 1-4 L3, 8-11 LF) et dix passes décisives et a terminé avec un double-double.
>
> Le Heat a compté 49 pour cent de tirs réussis et 20 lancers francs sur 29.  Mario Chalmers qui figurait dans l'alignement de départ a contribué un efficace 20 points avec six tirs réussis.  Luol Deng a fini avec 15 points.  Shawne Williams a enregistré 15 points (5-9 L, 3-5 L3, 2-2 LF) tout en ajoutant quatre passes décisives.  Dwyane Wade a marqué neuf points avec dix passes décisives.

> Les 76ers ont compté 52 pour cent de tirs réussis et 19 lancers francs sur 26 et ont subi 24 pertes de ballon.  Brandon Davies qui figurait dans l'alignement de départ a enregistré 18 points avec sept tirs réussis.  Luc Mbah a Moute a fini avec neuf points avec sept rebonds récupérés et trois passes décisives.  Malcolm Thomas a ajouté huit points en 19 minutes.  Alexey Shved a marqué six points avec six passes décisives.
>
> À venir pour le Heat, un match à domicile contre Toronto. Pour leur prochain match, les 76ers joueront à la maison contre les Rockets de Houston lundi.

more details

## 5.3.  Parallel generation of random data

This *nodata-to-text* example is nevertheless interesting because it illustrates the sentence modifications of *pyrealb* applied similarly to both languages. Random variations of sentence patterns are generated to create translation drill exercises.  This is a command-line version of a jsRealB web application which is more user-friendly to use, but the text generation algorithm is the same in both versions.

With *pyrealb* a sentence pattern can be parameterized using a lambda to change some of its words, inflections (number and tense) and even its structure by negating it, making it passive or interrogative. For example, the following definition in which the formal parameters are given *arbitrary* names but *easier* to remember in the context of the syntactical structure.

```
f = lambda n, child, eat, a, potato:\
       S(NP(D("the"), N(child).n(n)),
         VP(V(eat),
            NP(D(a), potato.n("p"))))
```

This definition can be used to produce different sentences, as shown in the following calls with the corresponding realizations. Lines 1-2 is a simple call changing terminals, lines 3-4 shows the negative future form of the sentence and lines 5-6 shows a negative and tag-interrogative form of the sentence in the past tense.

```
f("p","child","love","a",N("avocado")).realize()
=> 'The children love avocados. '
f("s","mother","cook","the",N("apple")).t("f").typ({"neg":True}).realize()
=> 'The mother will not cook the apples. '
f("s","uncle","eat","the",N("apple")).t("ps").typ({"neg":True,"int":"tag"}).realize()
=> 'The uncle did not eat the apples, did he? '
```

### 5.3.1. Translation drill exercises [code]

To create translation drill exercises, parallel sentence patterns are called with *equivalent* parameters and modified in the same way in the source and target languages. Thus two sentences structures can be created, one corresponding to the translation of the other. The realization of the source structure is shown to the user, while the tokens of the realization of the target structure are shuffled with some *distractor* words. The user types some tokens to create a translation that is compared with the expected realization of the target structure. Translation drills can be created in both translation directions by selecting which language is the source.

The following shows two interactions with the system.

```
1   Translate in English the sentences in French using some of the suggested words.
2   Type "end" to exit.
3   The child can love the watermelons.
4   ', adorer, ., manger, un, melons, eau, ', les, frère, peut, d, enfant, L
5   > L'enfant peut manger les melons d'eau.
6     L'enfant peut adorer les melons d'eau.:KO
7   Will the child love the watermelons?
8   ', enfant, t, -, adorera, un, ?, les, -, d, melons, ', soeur, eau, détester, L, il
9   > L'enfant adorera-t-il les melons d'eau?
10    L'enfant adorera-t-il les melons d'eau?:OK
11  The father will eat watermelons.
12  détester, père, enfant, eau, ', des, Le, d, ., mangera, le, melons
13  > end
```

The data for this demo is a list of *python* `dict`s that define the parameterized sentence structure for both languages with lists of alternatives for the parameters. The dict used in the previous example is shown below (lines 6-23). The parallel *pyreab* structures are created by two `lambda`s (lines 8-11 and 12-15). These functions are called by picking randomly in the lists of pairs of synonyms with their translation (lines 16-21) to assemble the actual parameters for the functions. Distractors are picked from the unchosen target values. Syntactical structures and the parameter values can differ in the source and target languages (see lines 20-21) but the lambdas must have the same number of *corresponding* parameters.

```
1   dets = [["un", "a"], ["le", "the"]]
2   numbers = [["s", "s"], ["p", "p"]]
3   relatives = [["père", "father"], ["frère", "brother"], ["soeur", "sister"],
4                ["tante", "aunt"]]
5
6   {"id": "F-02",
7    "level": 0,
8    "fr": lambda n, enfant, manger, un, pommeDT:
9           S(NP(D("le"), N(enfant).n(n)),
10              VP(V(manger),
11                 NP(D(un), pommeDT.n("p")))),
```

```
12      "en": lambda n, child, eat, a, potato:
13             S(NP(D("the"), N(child).n(n)),
14               VP(V(eat),
15                  NP(D(a), potato.n("p")))),
16      "params": [numbers,
17                 [["enfant", "child"], *relatives],
18                 [["manger", "eat"], ["adorer", "love"], ["détester", "hate"]],
19                 dets,
20                 [[lambda: NP(N("pomme"), PP(P("de"), N("terre"))), lambda: N("potato")],
21                  [lambda: NP(N("melon"), PP(P("de"), N("eau"))), lambda: N("watermelon")]]]
22    }
```

These parallel sentence patterns were created manually or adapted from an automatic transformation of the output of the Stanza parser described in (Lapalme 2021). Although for the moment there are only 8 sentence patterns, they can be realized in a variety of sentences when parameter, tense and sentence type variations are taken into account. As shown in the web demo, a teacher can also determine the level of difficulty of sentences going from simple affirmative sentences to complex passive interrogative forms.

## 6. What about GPT ?

GPT (Generative Pretrained Transformer) models have been used to generate many types of text, images and music. These billion-parameter systems, trained on terabytes of internet data, can be used to answer questions in a conversational manner.

We submitted the data for the examples of section 5 to the GPT instance of *Microsoft Copilot* (in November 2023) with *appropriate prompts* for text generation and translation (the appendix shows the complete transcript) which produced the following outputs.

[E2E challenge](#)

> "The Mill" is a pub situated in the riverside area, in proximity to "Raja Indian Cuisine".
> It serves English food and is known for its high price range. It is also family-friendly.

> "Le Mill" est un pub situé dans la zone riverside, à proximité de "Raja Indian Cuisine".
> Il sert de la nourriture anglaise et est connu pour sa gamme de prix élevée.
> Il est également adapté aux familles.

[WebNLG Challenge 2020](#)

> Alan Bean, born on March 15, 1932, in Wheeler, Texas, was selected by NASA in 1963. He was part of the Apollo 12 mission, which was operated by NASA. Alfred Worden served as the backup pilot for Apollo 12, and David Scott was the commander of the mission.

For these two examples, the results are excellent and comparable to the output of best systems that participated in these challenges. The output is more colloquial than the one produced by *pyrealb* and is well formatted. These types of input-output pairs, which appeared in the ACL Anthology, were probably included in the training set of the Transformer.

We also submitted the data shown above for the weather reports and the basketball summaries generation for which the results are impressive given the fact that the input was limited to a subset of the full data without any context.

[Weather reports](#)

> Here is the weather report based on the provided table:
>
> - From **5:00 to 15:00**, there is a **10%** chance of precipitation. The sky will be partially covered, with a sky cover value ranging from **2 to 8**. Temperatures will rise from **15°C** at 5:00 to a high of **28°C** at 14:00. The wind will be coming from the southwest with a speed of **10 km/h**.
>
> - From **15:00 to 18:00**, the chance of precipitation increases to **30%**, with possible showers or even a thunderstorm. The sky will be mostly covered, with a sky cover value of **8**. Temperatures will decrease to **25°C**. The wind will continue from the southwest but will increase to a speed of **20 km/h**.
>
> - From **18:00 to 20:00**, temperatures will further decrease to **23°C**. The wind will continue from the southwest at a speed of **20 km/h**.
>
> - Please note that the UV index will be **7.7** between **12:00 and 14:00**, which is considered high. It's recommended to wear protective clothing, a wide-brimmed hat, and UV-blocking sunglasses during this time.
>
> Please stay tuned for updates and stay safe! 😊

Copilot generated a Markdown formatted rendering grouping information by time interval ending with a smiley! In some cases, it combined time intervals with the corresponding temperatures. It also *hallucinated* a recommandation for the high UV index.

[Basketball summaries](#)

> During the game, a total of 96 points were scored. The team made 35 out of 67 field goal attempts, with 7 out of 23 successful three-point attempts. They also made 19 out of 26 free throw attempts.
>
> In the first quarter (Q1), the team scored 30 points, making 13 out of 21 field goal attempts and 3 out of 8 three-point attempts. They made 1 out of 2 free throw attempts.
>
> In the second quarter (Q2), the team scored 24 points, making 8 out of 15 field goal attempts. However, they didn't make any three-point attempts in this quarter. They made 8 out of 12 free throw attempts.
>
> In the third quarter (Q3), the team scored 29 points, making 10 out of 16 field goal attempts and 4 out of 8 three-point attempts. They made 5 out of 6 free throw attempts.
>
> In the fourth quarter (Q4), the team scored 13 points, making 4 out of 15 field goal attempts. They didn't make any three-point attempts in this quarter. They made 5 out of 6 free throw attempts.
>
> The team collected a total of 37 rebounds, with 4 of them being offensive rebounds. They also had 28 assists, 9 steals, and 10 blocks. However, they turned the ball over 24 times and committed 21 personal fouls.

In this case, given a box-score table with coded titles for columns and lines, Copilot managed to make sense of abbreviations such as *FG3M* and *FG3A*. This is better than us, because not being a basketball fan, we had no clue of what these codes meant. The text organisation is also good, because the results for the whole game are shown before the ones for each quarter.

Given the information for the four players scoring the most points, the result was this text which is factually correct, although a bit repetitive.

> Here is a summary of each player's performance:
>
> - **Tony Wroten**: He started the game and played for 33 minutes. He made 6 out of 11 field goal attempts, including 1 out of 4 three-point attempts. He also made 8 out of 11 free throw attempts. He collected 3 total rebounds and provided 10 assists. He had 1 steal, 1 block, 4 turnovers, and 1 personal foul. He scored a total of 21 points but had a negative impact on the team's score with a -11 plus/minus.
> - **Brandon Davies**: He came off the bench and played for 23 minutes. He made 7 out of 9 field goal attempts, including 1 out of 2 three-point attempts. He also made 3 out of 4 free throw attempts. He collected 3 total rebounds and had 3 steals. He had 3 turnovers and 3 personal fouls. He scored a total of 18 points and had a -1 plus/minus, indicating a nearly neutral impact on the team's score.
> - **Hollis Thompson**: He started the game and played for 32 minutes. He made 4 out of 8 field goal attempts, including 2 out of 5 three-point attempts. He did not attempt any free throws. He collected 1 total rebound and provided 2 assists. He had 3 blocks, 2 turnovers, and 2 personal fouls. He scored a total of 10 points but had a negative impact on the team's score with a -17 plus/minus.
> - **Henry Sims**: He started the game and played for 27 minutes. He made 4 out of 9 field goal attempts and made 1 out of 2 free throw attempts. He collected 4 total rebounds and provided 2 assists. He had 1 block, no turnovers, and 1 personal foul. He scored a total of 9 points and had a -10 plus/minus, indicating a negative impact on the team's score.

Of course, the results for the last two examples are difficult to compare with the *pyrealb* outputs because Copilot was given only a small subset of the data without any context. But even there it managed to realize excellent literal readings of the data.

In our experiments, each conversation turn took between 4 or 5 seconds of real time and probably costed [a few dozens of cents](). This is good enough for demoing, but this would probably be prohibitive in a production setting. This type of technology is just beginning and will surely be improved in terms in cost and execution time over the years, perhaps even months. *pyrealb* is *greener* as it runs very fast (one or two milliseconds per text) with limited memory requirements on a stock laptop without the need for a GPU. To be fair, we should take into account the development time as well. An NLG system like GPT has a development time that is amortized over all its possible applications, but it has a high cost of inference. For *pyrealb*, we have a long development time and almost none for inference.

Even though these *black box* systems show impressive results, they can be unpredictable (like humans!) During our short experiments we noticed that, when given the same prompt, Copilot did not always return exactly the same results, this can be problematic in some cases. The main advantage of a symbolic system is the control on the generated output for either the formulation or the phrasing. It also lend itself to interpretations, debugging and hardcoding of business rules. Reiter (2023) discusses some pros and cons of using large language models in a data-to-text context.

## 7. Conclusion

This document has shown *pyrealb* brought into play in different data-to-text contexts to convey information in French and English reliably through a convenient formalism that is familiar to linguists. The [*pyrealb* demo directory](#) shows other examples of features in a unilingual context, many of them can be run in either French or English but not in strictly bilingual mode. This approach could be extended to other languages provided that extensive lexicons and programs for implementing grammar rules are developed for them.

It is often argued that the drawback of a symbolic approach to generation is that sentence patterns must be developed manually by studying the corpus of reference texts. But we found this approach easier, faster and more fun than fine-tuning the parameters of a learning algorithm or tweaking prompts for an LLM. The fact the *pyrealb* caters automatically to conjugation, declension, agreements, elision, punctuation and formatting (HTML or not) greatly simplifies the building of sentence patterns to realize production quality texts. Patterns are defined at a relatively abstract level and can be realized in a variety of ways. It would be interesting and challenging to explore the possibility of *learning* sentence patterns from corpora.

The objective of this document was to demonstrate the organization of the generation process with *pyrealb*. Outside of a brief comparison with an instance of a GPT, It did not discuss any industrial exploitation or evaluation, which is an independent but important endeavor.

## 8. Acknowledgements

We thank Fabrizio Gotti and Ehud Reiter for interesting suggestions about a previous version of this paper.

- François Lareau, Florie Lambrey, Ieva Dubinskaite, Daniel Galarreta-Piquette, and Maryam Nejat. 2018. [GenDR: A Generic Deep Realizer with Complex Lexicalization](#). In *Proceedings of the Eleventh International Conference on Language Resources and Evaluation (LREC 2018)*, Miyazaki, Japan. European Language Resources Association (ELRA).
- Aarne Ranta, *Grammatical Framework: Programming with Multilingual Grammars*, CSLI Publications, Stanford, 2011, 340 pp, ISBN-10: 1-57586-626-9 (Paper), 1-57586-627-7 (Cloth).
- E. Reiter and R. Dale, Building natural language generation systems, Cambridge University Press, 2000.
- E. Reiter, [LLMs and Data-to-text](#), Ehud Reiter's Blog, June 29 2023.
- Ashish Upadhyay and Stewart Massie. 2022. [Content Type Profiling of Data-to-Text Generation Datasets](#). In *Proceedings of the 29th International Conference on Computational Linguistics*, pages 5770–5782, Gyeongju, Republic of Korea. International Committee on Computational Linguistics.
- P.-L. Vaudry and G. Lapalme, Adapting SimpleNLG for bilingual English - French realisation, in A. f. C. Linguistics, ed., 14th European Workshop on Natural Language Generation, Sofia, Bulgaria, 2013, pp. 183-187.

## 10.  Appendix: finding and exploring data

In a *real* application, the data is usually available although not always easy to collect, filter and curate. One should never underestimate the difficulty of getting high (or even moderate) quality data that often appears in various formats on different devices.

To develop new systems and compare their results with the state of the art, it can be useful to use existing data sets. For NLG, many of these data sets are associated with reference texts that can be exploited as targets for generation, corpus analysis or training machine learning systems. Most of these texts were authored by humans directly or by crowdsourcing, but care must be taken because [some of them were machine generated](#).

Here are some useful sources of NLG data:

- [Data sets for NLG](#) from the Natural Language Generation Portal
- [Data cards](#) from the GEM benchmark environment focusing on NLG evaluation
- [Hugging Face Datasets](#) is a source of well-organized datasets with an efficient *python* API. At the time of writing (October 2023),  there were more than 70,000 datasets (more than 1,500 being tagged *Text Generation*!), so it is best to search a data set by name.

Always take some time to **look at the data** and check the types and range of values. We have encountered JSON files in which all numeric values had been stored as strings; given that `"10" < "2"` is `True` and `"10"+"2"` equals `"102"`, this can give rise to unexpected results or sorting.

Here are some validation tools for text-based data, ignoring databases which have their own validation mechanisms:

- JSON: [JSON Schema](#), [JSON-rnc](#)
- XML: [Relax-NG](#)
- CSV: [CSV File Validator](#)

To explore and extract data:

- JSON: *jq* is very efficient for huge JSON files but its syntax can be a bit difficult to grasp at the start
    - [manual](#)
    - [excellent introduction](#) although now full of ads...
- XML: [XSLT](#), [ExamineXML](#)
- CSV : [awk](#)

## 11.  Appendix : Conversation with Microsoft Copilot

Transcript made in November 2023 selecting the *More Precise* mode

**Question**:
I want to write a completely factual restaurant description for a guide. It is important that the description be entirely factual, and be one or two-sentence long. This is what I know about the restaurant: name="The Mill" near="Raja Indian Cuisine" customer rating="unspecified" priceRange="high" area="riverside" familyFriendly="yes" food="English" eatType="pub"
Could you generate this restaurant description in English?

**Bing**:
"The Mill" is a family-friendly pub located in the riverside area, near "Raja Indian Cuisine". Known for its high-end English cuisine, it provides a warm and welcoming atmosphere for all patrons.

**Question**:
Could you rewrite this without any promotional tone, and be entirely factual?

**Bing**:
"The Mill" is a pub situated in the riverside area, in proximity to "Raja Indian Cuisine". It serves English food and is known for its high price range. It is also family-friendly.

**Question**:
Very good. Could you generate a French translation of your last answer?

**Bing**:
"Le Mill" est un pub situé dans la zone riverside, à proximité de "Raja Indian Cuisine". Il sert de la nourriture anglaise et est connu pour sa gamme de prix élevée. Il est également adapté aux familles.

**Question**:
Could you write a very short factual paragraph from this list of facts:
(Apollo_12 , backupPilot , Alfred_Worden) and
(Alan_Bean , mission , Apollo_12) and
(Apollo_12 , operator , NASA) and
(Apollo_12 , commander , David_Scott) and
(Alan_Bean , birthPlace , Wheeler,_Texas) and
(Alan_Bean , selectedByNasa , 1963) and
(Alan_Bean , birthDate , "1932-03-15").

lease don't add any additional facts to your answer. Stick to the facts expressed in my question.

**Bing**:

Alan Bean, born on March 15, 1932, in Wheeler, Texas, was selected by NASA in 1963. He was part of the Apollo 12 mission, which was operated by NASA. Alfred Worden served as the backup pilot for Apollo 12, and David Scott was the commander of the mission.

**Question**:

Very good. Could you generate a French translation of your last answer?

**Bing**:

Alan Bean, né le 15 mars 1932 à Wheeler, au Texas, a été sélectionné par la NASA en 1963. Il a fait partie de la mission Apollo 12, qui était opérée par la NASA. Alfred Worden a servi de pilote de secours pour Apollo 12, et David Scott était le commandant de la mission.

**Question**:

Given the following information triples about a university:

Visvesvaraya_Technological_University | city | Belgaum
Acharya_Institute_of_Technology | established | 2000
Acharya_Institute_of_Technology | motto | "Nurturing Excellence"
Acharya_Institute_of_Technology | state | Karnataka
Acharya_Institute_of_Technology | country | "India"
Acharya_Institute_of_Technology | campus | "In Soldevanahalli, Acharya Dr. Sarvapalli Radhakrishnan Road, Hessarghatta Main Road, Bangalore – 560090."
Acharya_Institute_of_Technology | affiliation | Visvesvaraya_Technological_University.

Describe this institution.

**Bing**:

The Acharya Institute of Technology is an educational institution located in India, specifically in the state of Karnataka. It was established in the year 2000 and is situated in Soldevanahalli, on Acharya Dr. Sarvapalli Radhakrishnan Road, Hessarghatta Main Road, Bangalore – 560090. The institute's motto is "Nurturing Excellence". It is affiliated with the Visvesvaraya Technological University, which is based in the city of Belgaum.

**Question**:

Given this weather information table:

| **tomorrow** | **( 6h,18h) fpto12-2018-07-18-2000-r1209c :: 2018-07-18 16:00:00** |
|---|---|
| precipitation-type | [15h,0h):[showers, [15h,0h):[thunderstorm]] |
| precipitation-probability | [5h,15h):[10], [15h,18h):[30] |
| sky-cover | [5h,11h):[2, 2], [11h,15h):[2, 8], [15h,18h):[8, 8] |
| temperatures : | [5h,8h):[15], [8h,11h):[23], [11h,14h):[28], [14h,17h):[25], [17h,20h):[23] |
| uv-index | [12h,14h):[7.7] |
| wind | [0h,12h):[sw, speed, 10], [12h,20h):[sw, speed, 20] |

can you produce a weather report.

**Bing**:Here is the weather report based on the provided table:

- From **5:00 to 15:00**, there is a **10%** chance of precipitation. The sky will be partially covered, with a sky cover value ranging from **2 to 8**. Temperatures will rise from **15°C** at 5:00 to a high of **28°C** at 14:00. The wind will be coming from the southwest with a speed of **10 km/h**.
- From **15:00 to 18:00**, the chance of precipitation increases to **30%**, with possible showers or even a thunderstorm. The sky will be mostly covered, with a sky cover value of **8**. Temperatures will decrease to **25°C**. The wind will continue from the southwest but will increase to a speed of **20 km/h**.
- From **18:00 to 20:00**, temperatures will further decrease to **23°C**. The wind will continue from the southwest at a speed of **20 km/h**.
- Please note that the UV index will be **7.7** between **12:00 and 14:00**, which is considered high. It's recommended to wear protective clothing, a wide-brimmed hat, and UV-blocking sunglasses during this time.

Please stay tuned for updates and stay safe! 😊

**Question**:
Given the following basketball box scores:

| Game | FGM | FGA | FG3M | FG3A | FTM | FTA | OREB | TREB | AST | STL | BLK | TOV | PF | PTS |
|---|---|---|---|---|---|---|---|---|---|---|---|---|---|---|
| Q1 | 13 | 21 | 3 | 8 | 1 | 2 | 2 | 10 | 10 | 2 | 5 | 6 | | 30 |
| Q2 | 8 | 15 | 0 | 2 | 8 | 12 | 1 | 11 | 7 | 1 | 1 | 4 | | 24 |
| Q3 | 10 | 16 | 4 | 8 | 5 | 6 | 1 | 10 | 9 | 2 | 2 | 6 | | 29 |
| Q4 | 4 | 15 | 0 | 5 | 5 | 6 | 0 | 6 | 2 | 4 | 2 | 8 | | 13 |
| game | 35 | 67 | 7 | 23 | 19 | 26 | 4 | 37 | 28 | 9 | 10 | 24 | 21 | 96 |

Describe what happened during this game.

**Bing**:
During the game, a total of 96 points were scored. The team made 35 out of 67 field goal attempts, with 7 out of 23 successful three-point attempts. They also made 19 out of 26 free throw attempts.

In the first quarter (Q1), the team scored 30 points, making 13 out of 21 field goal attempts and 3 out of 8 three-point attempts. They made 1 out of 2 free throw attempts.

In the second quarter (Q2), the team scored 24 points, making 8 out of 15 field goal attempts. However, they didn't make any three-point attempts in this quarter. They made 8 out of 12 free throw attempts.

In the third quarter (Q3), the team scored 29 points, making 10 out of 16 field goal attempts and 4 out of 8 three-point attempts. They made 5 out of 6 free throw attempts.

In the fourth quarter (Q4), the team scored 13 points, making 4 out of 15 field goal attempts. They didn't make any three-point attempts in this quarter. They made 5 out of 6 free throw attempts.

The team collected a total of 37 rebounds, with 4 of them being offensive rebounds. They also had 28 assists, 9 steals, and 10 blocks. However, they turned the ball over 24 times and committed 21 personal fouls.

**Question**:

Given the following basketball box scores for players:

| PLAYER | STRT | MIN | FGM | FGA | FG3M | FG3A | FTM | FTA | OREB | TREB | AST | STL | BLK | TOV | PF | PTS | +/- |
|---|---|---|---|---|---|---|---|---|---|---|---|---|---|---|---|---|---|
| Tony Wroten | True | 33 | 6 | 11 | 1 | 4 | 8 | 11 | 0 | 3 | 10 | 1 | 1 | 4 | 1 | 21 | -11 |
| Brandon Davies | False | 23 | 7 | 9 | 1 | 2 | 3 | 4 | 0 | 3 | 0 | 3 | 0 | 3 | 3 | 18 | -1 |
| Hollis Thompson | True | 32 | 4 | 8 | 2 | 5 | 0 | 0 | 0 | 1 | 2 | 0 | 3 | 2 | 2 | 10 | -17 |
| Henry Sims | True | 27 | 4 | 9 | 0 | 0 | 1 | 2 | 1 | 4 | 2 | 0 | 1 | 0 | 1 | 9 | -10 |

Describe the performance of each player.

**Bing**:

Here is a summary of each player's performance:

- **Tony Wroten**: He started the game and played for 33 minutes. He made 6 out of 11 field goal attempts, including 1 out of 4 three-point attempts. He also made 8 out of 11 free throw attempts. He collected 3 total rebounds and provided 10 assists. He had 1 steal, 1 block, 4 turnovers, and 1 personal foul. He scored a total of 21 points but had a negative impact on the team's score with a -11 plus/minus.

- **Brandon Davies**: He came off the bench and played for 23 minutes. He made 7 out of 9 field goal attempts, including 1 out of 2 three-point attempts. He also made 3 out of 4 free throw attempts. He collected 3 total rebounds and had 3 steals. He had 3 turnovers and 3 personal fouls. He scored a total of 18 points and had a -1 plus/minus, indicating a nearly neutral impact on the team's score.

- **Hollis Thompson**: He started the game and played for 32 minutes. He made 4 out of 8 field goal attempts, including 2 out of 5 three-point attempts. He did not attempt any free throws. He collected 1 total rebound and provided 2 assists. He had 3 blocks, 2 turnovers, and 2 personal fouls. He scored a total of 10 points but had a negative impact on the team's score with a -17 plus/minus.

- **Henry Sims**: He started the game and played for 27 minutes. He made 4 out of 9 field goal attempts and made 1 out of 2 free throw attempts. He collected 4 total rebounds and provided 2 assists. He had 1 block, no turnovers, and 1 personal foul. He scored a total of 9 points and had a -10 plus/minus, indicating a negative impact on the team's score.

**Question**:

Can you report only the outstanding performances of each player from your previous response.

**Bing**:

Sure, here are the outstanding performances of each player:

- **Tony Wroten**: He scored 21 points, made 6 out of 11 field goal attempts, and had 10 assists.
- **Brandon Davies**: He scored 18 points and made 7 out of 9 field goal attempts.
- **Hollis Thompson**: He made 3 blocks.
- **Henry Sims**: He had no turnovers.

These are the highlights based on the box scores provided. Please note that "outstanding" can be subjective and may vary depending on individual interpretation.